\def\year{2021}\relax
\documentclass[letterpaper]{article} 
\usepackage{aaai21}  
\usepackage{times}  
\usepackage{helvet} 
\usepackage{courier}  
\usepackage[hyphens]{url}  
\usepackage{graphicx} 
\usepackage{natbib}  
\usepackage{caption} 
\usepackage{multirow}
\usepackage{booktabs}
\usepackage{array,multirow}
\usepackage{tablefootnote}

\usepackage{amssymb}
\usepackage{pifont}
\newcommand{\cmark}{\ding{51}}%

\title{An Empirical Study of Cross-Lingual Transferability in Generative Dialogue State Tracker}
\author {
    Yen-Ting Lin,
    Yun-Nung Chen
    \\
}

\affiliations {
    National Taiwan University \\
    r08944064@csie.ntu.edu.tw, 
    y.v.chen@ieee.org
}
\begin{document}

\maketitle

\begin{abstract}
There has been a rapid development in data-driven task-oriented dialogue systems with the benefit of large-scale datasets.
However, the progress of dialogue systems in low-resource languages lags far behind due to the lack of high-quality data.
To advance the cross-lingual technology in building dialog systems, DSTC9 introduces the task of cross-lingual dialog state tracking, where we test the DST module in a low-resource language given the rich-resource training dataset.

This paper studies the transferability of a cross-lingual generative dialogue state tracking system using a multilingual pre-trained seq2seq model.
We experiment under different settings, including joint-training or pre-training on cross-lingual and cross-ontology datasets. 
We also find out the low cross-lingual transferability of our approaches and provides investigation and discussion. 
\end{abstract}
\section{Introduction}
Dialogue state tracking is one of the essential building blocks in the task-oriented dialogues system.
With the active research breakthrough in the data-driven task-oriented dialogue technology and the popularity of personal assistants in the market, the need for task-oriented dialogue systems capable of doing similar services in low-resource languages is expanding.
However, building a new dataset for task-oriented dialogue systems for low-resource language is even more laborious and costly.
It would be desirable to use existing data in a high-resource language to train models in low-resource languages.
Therefore, if cross-lingual transfer learning can be applied effectively and efficiently on dialogue state tracking, the development of task-oriented dialogue systems on low-resource languages can be accelerated.

The Ninth Dialog System Technology Challenge (DSTC9) Track2 \cite{Gunasekara2020} proposed a cross-lingual multi-domain dialogue state tracking task.
The main goal is to build a cross-lingual dialogue state tracker with a rich resource language training set and a small development set in the low resource language.
The organizers adopt MultiWOZ 2.1 \cite{eric2019multiwoz} and CrossWOZ \cite{Zhu2020a} as the dataset and provide the automatic translation of these two datasets for development.
In this paper's settings, our task is to build a cross-lingual dialogue state tracker in the settings of CrossWOZ-en, the English translation of CrossWOZ.
In the following, we will refer \textit{cross-lingual} datasets to datasets in different languages, such as MultiWOZ-zh and CrossWOZ-en, and \textit{cross-ontology} datasets to datasets with different ontologies, such as MultiWOZ-en and CrossWOZ-en.

The cross-lingual transfer learning claims to transfer knowledge across different languages.
However, in our experiments, we experience tremendous impediments in joint training on cross-lingual or even cross-ontology datasets.
To the best of our knowledge, all previous cross-lingual dialogue state trackers approach DST as a classification problem \cite{Mrksic2017a}\cite{Liu2019}, which does not guarantee the success of transferability on our generative dialogue state tracker.

The contributions of this paper are three-fold:
\begin{itemize}
\item This paper explores the cross-lingual generative dialogue state tracking system's transferability. 
\item This paper compares joint training and \textit{pre-train then finetune} method with cross-lingual and cross-ontology datasets.
\item This paper analyzes and open discussion on colossal performance drop when training with cross-lingual or cross-ontology datasets.
\end{itemize}

\section{Problem Formulation}
In this paper, we study the cross-lingual multi-domain dialogue state tracking task. 
Here we define the multi-domain dialogue state tracking problem and introduce the cross-lingual DST datasets.

\subsection{Multi-domain Dialogue State Tracking}
\begin{figure*}[h]
\centering
\includegraphics[width=0.7\linewidth]{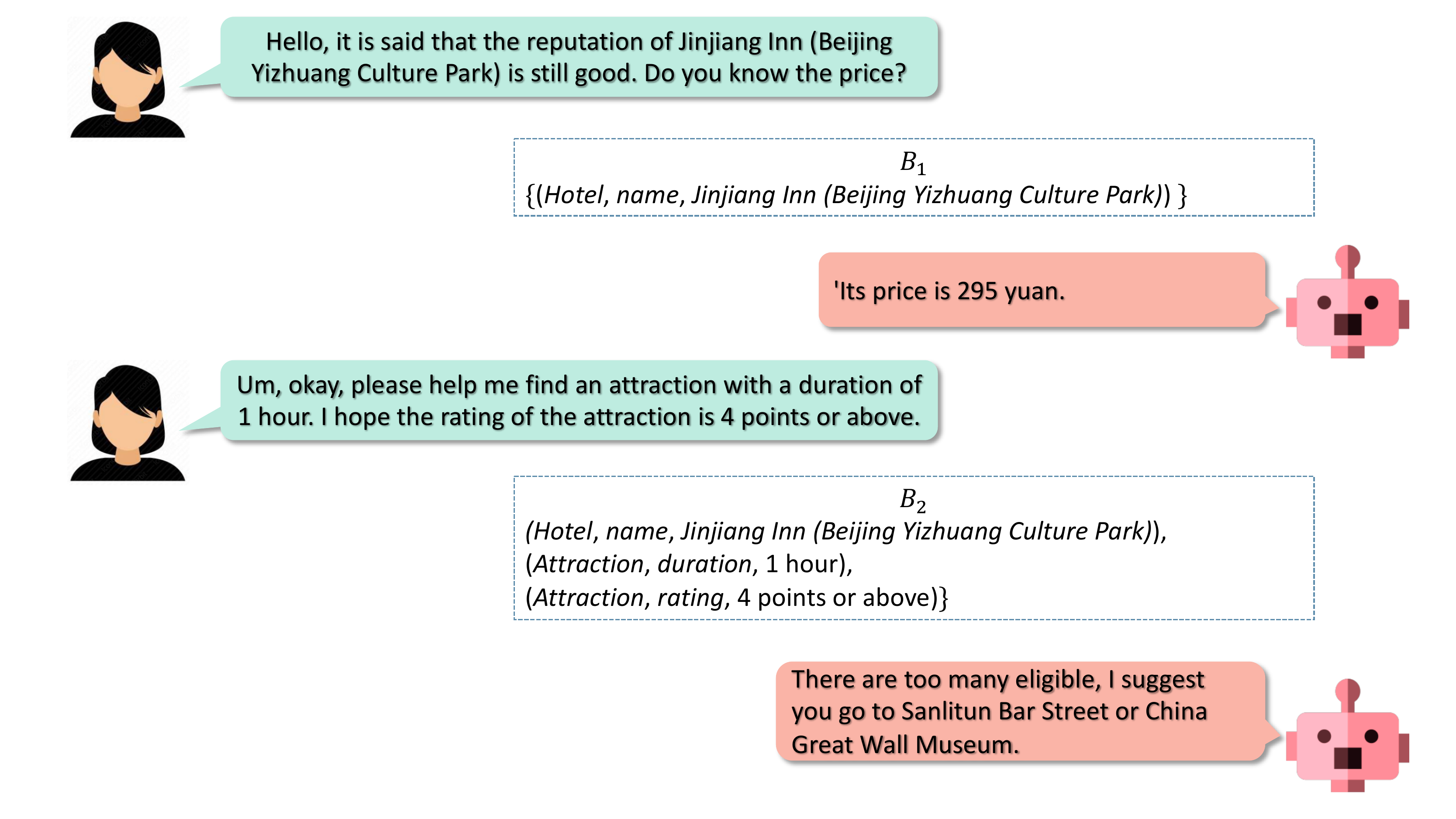}
\caption{Illustration of dialogue state tracking. The dialogue is sampled from CrossWOZ-en.}
\label{fig:dst}
\end{figure*}

The dialogue state in the multi-domain dialogue state tracking is a set of (\textit{domain, slot name, value}) triplets, where the domain indicates the service that the user is requesting, slot name represents the goal from the user, and value is the explicit constraint of the goal.
For dialogue states not mentioned in the dialogue context, we assign a \textit{null} value, \(\emptyset\), to the corresponding values.
For example, (\textit{Hotel, type, luxury}) summarizes one of the user's constraints of booking a luxury hotel, and (\textit{Attraction, fee, 20 yuan or less}) means the user wants to find a tourist attraction with a ticket price equal to or lower than 20 dollars. 
An example is presented in Figure\ref{fig:dst}.

Our task is to predict the dialogue state at the \(t^{th}\) turn, \(\mathcal{B}_t=\{(\mathcal{D}^i, \mathcal{S}^i, \mathcal{V}^i ) \,|\,1 \leq i \leq I \}\) where \(I\) is the number of states to be tracked, given the historical dialogue context until now, defined as \(\mathcal{C}_{t}=\{\mathcal{U}_1, \mathcal{R}_1, \mathcal{U}_2, \mathcal{R}_2,\dots, \mathcal{R}_{t-1}, \mathcal{U}_t\}\) where \(\mathcal{U}_i\) and \(\mathcal{R}_i\) is the user utterance and system response, respectively, at the \(i^{th}\) turn.

\subsection{Dataset}
MultiWOZ is the task-oriented dataset often used as the benchmark dataset for task-oriented dialogue system tasks, including dialogue state tracking, dialogue policy optimization, and NLG. MultiWOZ 2.1 is a cleaner version of the previous counterpart with more than 30\%  updates in dialogue state annotations.
CrossWOZ is a Chinese multi-domain task-oriented dataset with more than 6,000 dialogues, five domains, and 72 slots. 
Both of the above datasets collects human-to-human dialogues in Wizard-of-Oz settings.
Table \ref{tab:dataset} lists the details of the dataset.

In DSTC9 Track 2, the organizers translate MultiWOZ and CrossWOZ into Chinese and English, respectively, and we refer the translated version of MultiWOZ and CrossWOZ as MultiWOZ-zh and CrossWOZ-en, respectively.
The public and private test of CrossWOZ-en in DSTC9 has 250 dialogues, but only the public test set has annotations. Therefore, we use the public one as the test set in our experiments.

\begin{table}[h]
\centering
\renewcommand{\arraystretch}{1.2}
\begin{tabular}{|l||c|c|}
\hline
\textbf{Metric}         & \textbf{MultiWOZ} & \textbf{CrossWOZ}     \\ 
\hline
\textbf{Language}       & English           & Chinese (Simplified)  \\
\textbf{\# Dialogues}   & ~~~8,438                           & ~~5,012                                     \\
\textbf{Total \# turns} & 113,556                             & 84,692                                    \\
\# \textbf{Domains}     & 7                           & 5                                         \\
\# \textbf{Slots}       & 24                                  &72                                        \\
\# \textbf{Values}      & 4,510                               & 7,871                                     \\
\hline
\end{tabular}
\caption{Statistics of MultiWOZ and CrossWOZ. Note that the translated version of these two datasets have the same metrics}
\label{tab:dataset}
\end{table}

\section{Related Work}
\subsection{Dialogue State Tracker}
Traditionally, dialogue state tracking depends on fixed vocabulary approaches where retrieval-based models ranks slot candidates from a given slot ontology. \cite{ramadan-etal-2018-large}\cite{Lee2019}\cite{Shan2020}
However, recent research efforts in DST have moved towards generation-based approaches where the models generate slot value given the dialogue history.
\cite{Wu2019} proposed a generative multi-domain DST model with a copy mechanism which ensures the capability to generate unseen slot values.
\cite{Kim2019} introduced a selectively overwriting mechanism, a memory-based approach to increase efficiency in training and inference.
\cite{Le2020} adopted a non-autoregressive architecture to model potential dependencies among (domain, slot) pairs and reduce real-time DST latency significantly. 
\cite{Hosseini-Asl2020} took advantage of the powerful generation ability of large-scale auto-regressive language model and formulated the DST problem as a casual language modeling problem.

\subsection{Multilingual Transfer Learning in Task-oriented Dialogue}
\cite{schuster-etal-2019-cross-lingual} introduced a multilingual multi-domain NLU dataset. 
\cite{Mrksic2017a} annotated two additional languages to WOZ 2.0 \cite{Mrksic2017} and 
\cite{Liu2019} proposed a mixed-language training for cross-lingual NLU and DST tasks. 
Noted that all previous multilingual DST methods modeled the dialogue state tracking task as a classification problem. \cite{Mrksic2017a}\cite{Liu2019}

\section{Methods}
This paper considers the multi-domain dialogue state track-ing as a sequence generation task by adopting a sequence-to-sequence framework.

\subsection{Architecture}
Following \cite{Liu2020}, we use the  sequence-to-sequence Transformer architecture \cite{vaswani2017attention} with 12 layers in each encoder and decoder.
We denote \textit{seq2seq} as our model in the following.

\subsection{DST as Sequence Generation}
The input sequence is composed of the concatenation of dialogue context  \(\mathbf{x^t}=\{\mathcal{U}_1; \mathcal{R}_1; \mathcal{U}_2; \mathcal{R}_2;\dots; \mathcal{R}_{t-1}; \mathcal{U}_t\}\) where \(;\) denote the concatenation of texts.

For the target dialogue state, we only consider the slots where the values are non-empty.
The target sequence is consist of the concatenation of the (\textit{domain, slot, value}) triplets with a non-empty \textit{value}, \(\mathbf{y^t}=\{\mathcal{D}^i; \mathcal{S}^i; \mathcal{V}^i | 1\leq i \leq I \wedge \mathcal{S}^i \neq \emptyset \} \).
\[\mathbf{\hat{y}^t}=seq2seq(\mathbf{x^t})\]
We fix the order of the  (\textit{domain, slot name, value}) triplets for consistency.

The training objective is to minimize the cross-entropy loss between the ground truth sequence \(\mathbf{y^t}\) and the predicted sequence \(\mathbf{\hat{y}^t}\).

\subsection{Post-processing}
The predicted sequence \(\mathbf{\hat{y}^t}\) is then parsed by heuristic rules to construct \(\hat{\mathcal{B}_t}=\{\mathcal{D}^i; \mathcal{S}^i; \hat{\mathcal{V}}^i | 1\leq i \leq I\}\).

By utilizing the possible values of slots in the ontology, for predicted slot values \(\hat{\mathcal{V}}\) that do not appears in the ontology, we choose the one with the best match to our predicted value. \footnote{This is implemented by \textit{difflib.get\_close\_matches} in Python}

\section{Experiments}
In the following section, we describe evaluation metrics, experiment setting and introduce experimental results.

\subsection{Evaluation Metrics}
We use joint goal accuracy and slot F1 as our metrics to evaluate our dialogue state tracking system.
\begin{itemize}
\item Joint Goal Accuracy: The proportion of dialogue turns where predicted dialogue states match entirely to the ground truth dialogue states.
\item Slot F1: The macro-averaged F1 score for all slots in each turn.
\end{itemize}

\subsection{Experiments Settings}
We want to examine how different settings affect the performance of the target low-resource dataset: CrossWOZ-en.\footnote{In our experimental circumstance, English is the low-resource language since the original language of CrossWOZ is Chinese.}
We will conduct our experiments in the settings below.
\begin{itemize}
\item \textbf{Direct Fine-tuning}
\item \textbf{Cross-Lingual Training (CLT)}
\item \textbf{Cross-Ontology Training (COT)}
\item \textbf{Cross-Lingual Cross-Ontology Training (CL/COT)}
\item \textbf{Cross-Lingual Pre-Training (CLPT)}
\item \textbf{Cross-Ontology Pre-Training (COPT)}
\item \textbf{Cross-Lingual Cross-Ontology Pre-Training (CL/COPT)}
\end{itemize}
Table \ref{tab:results} and \ref{tab:pretrain_results} show the datasets for training and pre-training in different settings.
For experiments with pre-training, all models are pre-trained on the pre-training dataset and then fine-tuned on CrossWOZ-en.

The baseline model provided by DSTC9 is SUMBT \cite{Lee2019}, the ontology-based model trained on CrossWOZ-en.

\subsection{Multilingual Denoising Pre-training}
All of our models initialize from \textit{mBART25}. \cite{Liu2020}
\textit{mBART25} is trained with denoising auto-encoding task on mono-lingual data in 25 languages, including English and Simplified Chinese.
\cite{Liu2020} shows pre-training of denoising autoencoding on multiple languages improves the performance on low resource machine translation.
We hope using \textit{mBART25} as initial weights would improve the cross-lingual transferability.

\subsection{Implementation Details}
In all experiments, the models are optimized with AdamW \cite{Loshchilov2017} with learning rate set to $1e^{-4}$ for 4 epochs. 
The best model is selected from the validation loss and is used for testing.

During training, the decoder part of our model is trained in the teacher forcing fashion \cite{williams1989learning}.
Greedy decoding \cite{vinyals2015neural} is applied when inference.
Following mBART \cite{Liu2020}, we use sentencespiece tokenizer.
For GPU memory constraints, source sequences longer than 512 tokens are truncated at the front and target sequences longer than 256 tokens are truncated at the back.

The models are implemented in Transformers \cite{Wolf2019HuggingFacesTS}, PyTorch \cite{NEURIPS2019_9015} and PyTorch Lightning \cite{falcon2019pytorch}.

\section{Results and Discussion}
The results for all experiment settings are shown in Table~\ref{tab:results} and \ref{tab:pretrain_results}.

\begin{table*}[t]
\centering
\renewcommand{\arraystretch}{1.2}
\begin{tabular}{ | c || c | c | c | c || c | c | }
\hline
\multirow{3}{*}{\textbf{Experiment}} & \multicolumn{4}{c||}{\textbf{Training Data}} & \multirow{3}{*}{\textbf{JGA}} & \multirow{3}{*}{\textbf{SF1}} \\ 
 \cline{2-5}
 & \multicolumn{2}{c|}{\textbf{MultiWOZ}} & \multicolumn{2}{c||}{\textbf{CrossWOZ}} & & \\
 \cline{2-5}
 & \textbf{en} & \textbf{zh} & \textbf{en} & \textbf{zh} & & \\
 \hline
 \textbf{Baseline} &  &  &  \cmark &  &  ~~7.41 & ~~55.27*\\
 \hline
 \hline
 \textbf{Direct Fine-tuning} &  &  & \cmark &  & \textbf{16.82} & \textbf{66.35} \\
\textbf{CL/COT}  & \cmark & \cmark & \cmark & \cmark & ~~4.10 & 26.50 \\
\textbf{COT} & \cmark &  & \cmark &  & ~~0.95 & 19.60 \\
\textbf{CLT} &  &  & \cmark & \cmark & ~~0.53 & 13.45 \\
 \hline
\end{tabular}
\caption{Experimental results on CrossWOZ-en with different training data (\%). 
*: This slot f1 is averaged over both the public and private test dialogues. 
JGA: Joint Goal Accuracy. SF1:  Slot F1.}\smallskip
\label{tab:results}
\end{table*}

\begin{table*}[t]
\centering
\renewcommand{\arraystretch}{1.2}
\begin{tabular}{ | c || c | c | c | c || c | c | }
\hline
\multirow{4}{*}{\textbf{Experiment}} & \multicolumn{4}{c||}{\textbf{Pre-training Data}} & \multirow{3}{*}{\textbf{JGA}} & \multirow{3}{*}{\textbf{SF1}} \\ 
 \cline{2-5}
 & \multicolumn{2}{c|}{\textbf{MultiWOZ}} & \multicolumn{2}{c||}{\textbf{CrossWOZ}} & & \\
 \cline{2-5}
 & \textbf{en} & \textbf{zh} & \textbf{en} & \textbf{zh} & & \\
 \hline
\textbf{Direct Fine-tuning} & & & & &  \textbf{16.82} & \textbf{66.35}\\
\hline
\textbf{CL/COPT} & \cmark & \cmark &  &  & ~~5.94 & 38.36 \\
\textbf{COPT} & \cmark &  &  & & ~~2.52 & 27.01 \\
\textbf{CLPT} &  &  &  & \cmark & ~~0.11 & 15.01 \\
 \hline
\end{tabular}
\caption{Experimental results on CrossWOZ-en with pre-training (\%).}
\label{tab:pretrain_results}
\end{table*}

\subsection{Additional Training Data Cause Degeneration}
 \textit{Direct Fine-tuning} significantly outperforms other settings, including the official baseline.
We assume English and Chinese data with the same ontology to train the mBART would bridge the gap between the two languages and increase the performance.
However, in \textit{Cross-Lingual Training}, training on English and Chinese version of CrossWOZ leads to catastrophic performance on CrossWOZ-en.

In the \textit{Cross-Ontology Training} where combine two data in the same language.
However, with different ontologies, the performance marginally increases from \textit{Cross-Lingual Training}, which shows more extensive mono-lingual data with the unmatched domain, slots, and ontology confuses the model during inference.
In the \textit{Cross-Lingual Cross-Ontology Training}, we collect all four datasets for training, and the performance is still far from \textit{Direct Fine-tuning}.

In conclusion, additional data deteriorate the performance on CrossWOZ-en even whether the language or ontology matches or not.
 
 \subsection{Does "First Pre-training, then fine-tuning" Help?}
We hypothesize that training with additional data causes performance degeneration, and therefore one possible improvement could be first pre-training the model on cross-lingual / cross-ontology data and then fine-tuning on the target dataset CrossWOZ-en.
Table \ref{tab:pretrain_results} shows the results.

By comparing \textit{COPT} to \textit{COT} and \textit{CL/COPT } to \textit{CL/COP}, the relative performance gain by over 37\% with regards to slot F1.
"Pre-training, fine-tuning" framework may partially alleviate the problem of catastrophic performance drop in joint training.

\subsection{Domain Performance Difference across Experiment Settings?}
\begin{figure}[h]
\centering
\includegraphics[width=1\linewidth]{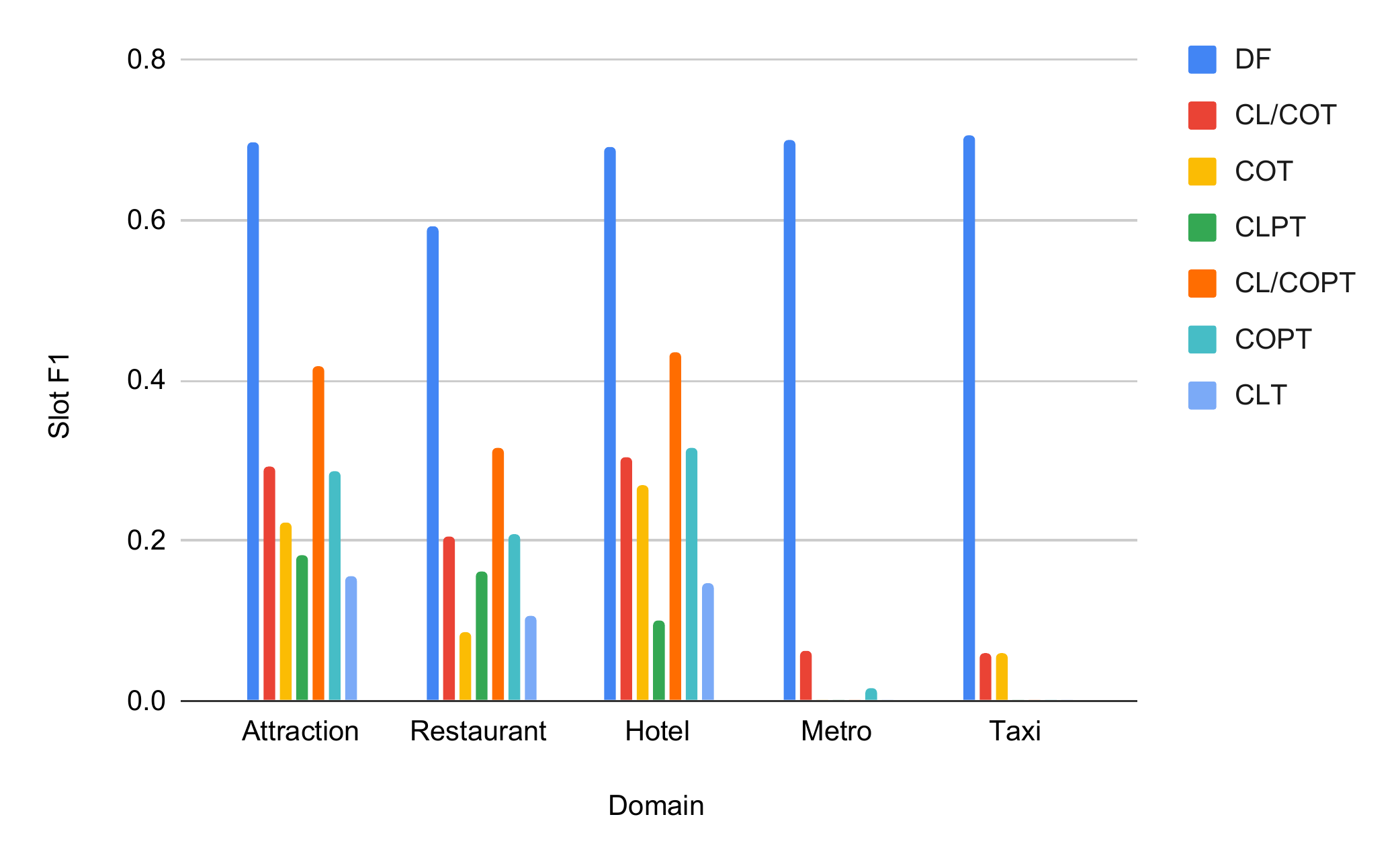}
\caption{Slot F1 across 5 domains in CrossWOZ-en in different settings.}
\label{fig:sf1}
\end{figure}

This section further investigates the cause of the performance decrease by comparing the slot F1 of different models across five domains in Figure \ref{fig:sf1}.

Generally speaking, in attraction, restaurant, and hotel domains, "pre-train then fine-tune" methods beat their "joint training" counterparts by an observable margin.
By contrast, in metro and taxi domains, despite poor performance among all, "joint training" settings beat their"pre-train then fine-tune" counterparts.

The only two trackable slots in the metro and taxi domain, "from" and "to," usually take the address or name of buildings, are highly non-transferable across datasets.
We conjecture that pretraining on cross-lingual or cross-ontology datasets does not help or even hurt those non-transferable slots.

\section{Conclusion}
In this paper, we build a cross-lingual multi-domain generative dialogue state tracker with multilingual seq2seq to test on CrossWOZ-en and investigate our tracker's transferability under different training settings.
We find that jointly trained the dialogue state tracker on cross-lingual or cross-ontology data degenerates the performance. 
\textit{Pre-training on cross-lingual or cross-ontology data, then fine-tuning} framework may alleviate the problem, and we find empirically evidence on relative improvement in slot F1.
A finding from the domain performance shift is that performance on some non-transferable slots, such as \textit{name}, \textit{from}, \textit{to}, may be limited by the previous pretraining approach.
A future research direction would investigate why such a significant performance declines in joint training and tries to bridge it.

\bibliography{LaTeX/mybib,LaTeX/mendeley}
\end{document}